# Model Choices Influence Attributive Word Associations: A Semi-supervised Analysis of Static Word Embeddings

Geetanjali Bihani
*Computer and Information Technology*
*Purdue University*
West Lafayette, IN, USA
gbihani@purdue.edu

Julia Taylor Rayz
*Computer and Information Technology*
*Purdue University*
West Lafayette, IN, USA
jtaylor1@purdue.edu

*Abstract*—Static word embeddings encode word associations, extensively utilized in downstream NLP tasks. Although prior studies have discussed the nature of such word associations in terms of biases and lexical regularities captured, the variation in word associations based on the embedding training procedure remains in obscurity. This work aims to address this gap by assessing attributive word associations across five different static word embedding architectures, analyzing the impact of the choice of the model architecture, context learning flavor and training corpora. Our approach utilizes a semi-supervised clustering method to cluster annotated proper nouns and adjectives, based on their word embedding features, revealing underlying attributive word associations formed in the embedding space, without introducing any confirmation bias. Our results reveal that the choice of the context learning flavor during embedding training (CBOW vs skip-gram) impacts the word association distinguishability and word embeddings' sensitivity to deviations in the training corpora. Moreover, it is empirically shown that even when trained over the same corpora, there is significant inter-model disparity and intra-model similarity in the encoded word associations across different word embedding models, portraying specific patterns in the way the embedding space is created for each embedding architecture.

## I. INTRODUCTION

Static word embedding models, based on the theory of distributional semantics, are increasingly being employed to perform downstream NLP tasks [1]–[3]. These models infer the meaning of words using the distribution of linguistic contexts they appear in [4], encoding word representations within high dimensional embedding spaces by learning semantic and syntactic regularities in language [5]–[7]. Due to the underlying distributional hypothesis utilized within the model architecture, these models have shown susceptibility to encoding useful as well as undesirable word associations present within natural language data [8]–[11]. This raises questions regarding the validity of the application of these models in tasks affecting real-world decisions such as web retrieval [12], user profiling [13], recommendation engines [14], etc.

The examination of word associations encoded within such word embedding spaces has been limited to finding the evidence of undesirable associations and their removal [8], [9], [11]. Since word embedding models do not explicitly encode linguistic relations, it becomes difficult to analyze why a particular embedding space is generated and why certain word associations implicitly develop, impacting downstream NLP tasks [15]–[17]. Thus, it is crucial to assess the role played by the model architecture, training corpora and context learning process in encoding associations between words within embedding spaces.

As a first step towards addressing this concern, our work presents a methodology to assess how model architecture and corpus choices influence attributive word associations developed across different word embedding models. We choose to analyze attributive word associations based on prior work examining associations between word embeddings for target and attribute words [17], [18]. Our analysis focuses on annotated proper nouns as neutral words and adjectives as attribute words, since they describe some quality of a noun, to generate attributive word associations. A semi-supervised approach is introduced to eliminate confirmation bias when choosing words and analyzing respective relationships. The analysis is conducted for five different word embedding frameworks i.e. word2vec (CBOW & skip-gram) [18], GloVe [19], fastText (CBOW & skip-gram) [20].

## II. WORD EMBEDDING MODELS

Word embeddings are distributed word representations mapping a given word to a vector in a low-dimensional space. This paper analyzes three static word embedding frameworks: word2vec [18], GloVe [19] and fastText [20] and the two different context learning approaches used to train word2vec and fastText embeddings, i.e. CBOW and skip-gram.

The word2vec framework [18] utilizes a shallow neural network architecture to create embeddings, addressing various gaps in the prior word representation methods that incurred high computational costs for large amounts of data. The method reduces the size of word embedding space and densely packs more information within each embedding, pointing at semantic and syntactic regularities [7]. However, while preserving distances between two words, this framework discards the statistical properties of the corpus.

The GloVe framework [19] combines global matrix factorization with local context window methods to learn and create word embeddings. Unlike word2vec, which predicts a given word using neural-networks, GloVe uses a frequency-based approach to count the number of times a word has appeared in a context and creates a co-occurrence matrix based on this information. It was conceptualized as an approach to address the lack of transparency regarding semantic and syntactic regularities in prior embedding frameworks.

FastText [20] was proposed as an extension of the word2vec model, but with a different treatment of each word during the embedding creation process. Contrasting from word2vec and GloVe, it takes the morphology of words into account and treats each word as a bag of character n-grams. Each character n-gram has a vector representation, and a word composed of multiple character n-grams has a vector representation of the sum of its component character n-grams.

The two context learning approaches that are used to train word2vec and fastText embeddings differ in their treatment of word-context relation. In CBOW i.e. continuous bag of words, the network tries to predict which word is most likely to occur given its context, while in skip-gram, the network tries to predict the neighboring words (context) which are most likely to occur, given the current word.

## III. METHODOLOGY

We develop a corpus-adapting word association evaluation approach, assessing the impact of model architecture and input text corpora on developed word associations. By training the same model on different inputs, we evaluated whether the word associations captured pertained to the respective model, regardless of the variations in the input corpora. By training different models using the same text corpora, we evaluated how word associations captured can differ across models, although stemming from the same input corpora.

The process began by identifying neutral and attribute words from the raw text corpus. The text was lowercased, special and numeric characters removed, and words tokenized, finally sent as an input to the word embedding model(s) to be converted to high dimensional vectors. These word vectors were used to cluster neutral and attribute words together, using an agglomerative hierarchical clustering approach. Inter-cluster and intra-cluster associations were assessed to provide information regarding the evolution of word associations over changing word embedding models and text corpora.

### A. Corpus selection

The Corpus of Historical American English [21] was used as the text input for word embedding training and analysis. This corpus captures a wide range of linguistic regularities from texts spanning twenty decades (1810s to 2000s), enabling the observation of variations in lexis, morphology, semantic and syntactic structures. Thus, this corpus acted as a large and highly granular dataset of words occurring within different contexts, enabling the analysis of the evolution of word associations over a wide range of linguistic variation across twenty decades.

### B. Neutral and Attribute Word Tagging

Since prior bias identification works utilize proper nouns as neutral target words, due to their susceptibility to capturing wanted and unwanted word associations [11], [22], this study utilized proper nouns as neutral words. Adjectives being attributive in nature when considered alongside proper nouns [23], have been utilized as attribute words.

Neutral and attribute word selection has been automated to reduce the inclusion of researchers' biases during word selection for word association evaluation, and to address implicit linguistic variations in the input corpora, not covered by human intuition. Words automatically tagged as either proper nouns or adjectives, using the Stanza NLP toolkit [24] part-of-speech tagger, have been chosen.

### C. Training Word Embeddings

The raw text corpora spanning twenty decades, served as the input to create word embeddings using the five word embedding frameworks: word2vec (CBOW & skip-gram), GloVe, fastText (CBOW & skip-gram). In order to keep the comparison between different word embedding models consistent, the hyperparameters were kept uniform: The window size was kept at 5 and the learning rate at 0.025, for all the five models. It should be noted that GloVe is disadvantaged when using the same number of epochs as word2vec and fastText.

### D. Neutral-Attribute Word Clustering

Word embeddings capture word relatedness for words appearing in similar contexts, encoding them such that they exist closer within the vector space. Hence, it is expected that the word embeddings of a proper noun and an adjective, appearing within the same context will lie closer as compared to the word embeddings of a proper noun and an adjective not employed within the same context.

Thus, this relatedness between neutral and attribute words within the word embedding space was explored by evaluating the tendency of these words to cluster together. Agglomerative hierarchical clustering was utilized to minimize human intervention when creating proper noun and adjective word clusters and providing a deterministic and more flexible clustering mechanism [25]. At each merging step, the merging cost of combining a pair of clusters was accounted for by using Ward linkage as the clustering metric, uncovering clusters that might not have been round or similar in diameter. Cosine distance was used to compute the proximity between word embeddings, bounded by the range [0,1]. Thus, the more related two words were, the smaller the distance between their vector representations leading to a higher tendency of them being in the same cluster, and vice versa.

The given corpus provided a wide range of variability in word usage, implicit semantic themes, syntactic regularities and overall lexical richness, leading to variations in the optimal number of word clusters that needed to be created across the corpus from each decade. As computational heuristic methods could not identify a clear preference of a cluster number, in order to define the optimal number of word clusters for every decade, this paper utilized the theories that inform distinctions between adjectives. Prior literature shows that adjectives tend to have semantic orientations and gradeability attached to them [26]. Semantic orientations, among other things, refer to an adjective being neutral, positive or negative. Gradeability on the other hand refers to an adjective dealing with comparative constructs, such as small versus large. The root morphemes of some adjectives can also be traced to emotions [27]. For example, words like 'fearful' can be traced back to the root morpheme 'fear', which can be traced to the emotion of fear itself. Thus, we made an assumption that non-neutral adjectives can be clustered by emotions.

In order to select the number of emotions that would correspond to the number of clusters, we utilized the Plutchik's wheel of emotions [28], a classification approach to distinguish emotions into 8 prototype emotions: joy, sadness, anger, fear, trust, disgust, surprise and anticipation, stating that the eight prototype emotions create all other emotions, which are mixed or derivative states, occurring as combinations or compounds of these primary emotions. Thus, the emotions represented by most adjectives should ideally be traced back to these 8 prototype emotion themes, defining it as the optimal number of distinguishing clusters, capturing distinctions across adjectives based on the emotion theme they represented.

The neutral-attribute (noun-adjective) word clusters and respective word associations were evaluated by assessing the quality and distribution of created clusters using inter-cluster and intra-cluster metrics. The inter-cluster analysis was done using Dunn's Index (eq. 1), which evaluated the quality of word clusters being formed by assessing the spread of word embedding clusters across the vector space. Ranging from 0 to ∞, this metric provided information regarding the quality of clusters being formed. This metric assessed cluster validity and aimed to identify tightly knit and well-separated clusters [29].

$$\text{Dunn's Index} = \frac{\text{Minimum inter cluster distance}}{\text{Maximum cluster diameter}} \quad (1)$$

For K given clusters, Dunn's Index was calculated as Du(K) as follows;

$$Du(K) = \min_{i=1,\ldots,K} \left( \min_{j=i+1,\ldots,K} \left( \frac{D(C_i, C_j)}{\max_{l=1,\ldots,K} \text{diam}(C_l)} \right) \right) \quad (2)$$

$$D(C_i, C_j) = \min_{x \in C_i, y \in C_j} D(x, y) \quad (3)$$

$$\text{diam}(C_i) = \max_{x,y \in C_i} D(x, y) \quad (4)$$

Dunn's Index gives us an outlook regarding which word embedding models generate the most distinctive word associations. Here, the distinctiveness of a word association captured by a word embedding model was attributed to the model's sensitivity towards capturing the similarity and dissimilarity between words. Hence, word clusters with a higher Dunn's Index pointed towards compact and well separated clusters, which in turn signified distinctive or distinguishable word associations, i.e. the model was more sensitive towards capturing similarities and dissimilarities between words, as compared to word clusters with a lower Dunn's Index.

For intra-cluster analysis, the clustering quality was assessed using two metrics. The first metric evaluates the distribution of words across clusters, comparing how uniformly the words have been grouped across clusters, for each word embedding model. Thus, if a word embedding model created balanced clusters, the proportion of words across each cluster remained fairly equal, as compared to a word embedding model creating unbalanced clusters. Over the same training corpora, the distribution of words across clusters revealed the differences in which words were associated with each other, depending on different word embedding frameworks.

The second metric averages Jaccard Similarity between clusters for each word embedding model pair, comparing the similarities and differences between the clustering patterns across different word embedding models (eq. 5).

$$J(A,B) = \frac{|A \cap B|}{|A \cup B|} = \frac{|A \cap B|}{|A| + |B| - |A \cap B|} \quad (5)$$

Thus, clustering done using word embedding models $W_A$ and $W_B$ is compared using Jaccard Similarity, extended to the respective sets of clusters, $C_A$: [$C_{A1}, \ldots, C_{A8}$] and $C_B$: [$C_{B1}, \ldots, C_{B8}$], calculated for all 64 combinations (eq. 6):

$$J(C_A, C_B) = \frac{J(C_{A1}, C_{B1}) + \ldots + J(C_{A8}, C_{B8})}{64} \quad (6)$$

The average Jaccard Similarity between clusters created by word embedding models $W_A$ and $W_B$, over all the twenty decades (1810s – 2000s) was calculated (eq. 7):

$$\text{Avg. } J(C_A, C_B)_{20} = \frac{J(C_A, C_B)_{1810s} + \ldots + J(C_A, C_B)_{2000s}}{20} \quad (7)$$

Thus, clusters created for every word embedding model were compared with the other four word embedding model clusters, trained across the same text corpora, revealing the similarities or differences between the clusters created using embeddings from different word embedding architectures. If word embedding model $W_A$ showed higher Jaccard similarity with model $W_B$ as compared to model $W_C$, it signified that the clusters created by $W_A$ and $W_B$ were more similar, i.e. the word associations being captured by both the models were more similar, as compared to $W_C$.

IV. RESULTS

The systematic evaluation of deviations in word associations performed over different word embedding models and different text corpora is described in TABLE I.

TABLE I. WORD ASSOCIATION EVALUATION FRAMEWORK

| Evaluation | Fixed | Varying | Observations |
|---|---|---|---|
| Evaluation I | Text corpus | WE model | • Dunn's Index (DI)<br>• Distribution of word within clusters<br>• Jaccard Similarity |
| Evaluation II | WE model | Text corpus | • Dunn's Index (DI)<br>• Corpus Size |

A. *Variation in word associations over varying WE models*

   *1) Variation in Dunn's Index*

Variations in Dunn's Index (DI) of clusters created using the five word embedding models, over a period of twenty decades, are shown in Fig. 1. Average Dunn's Index and standard deviation over twenty decades is also reported in TABLE II.

The results show that word2vec (skip-gram) embeddings portray the highest Dunn's Index, creating compact and well separated clusters, across all twenty decades of text data. Thus, words embedded using word2vec (skip-gram) model formed more distinctive word associations than the word associations formed using the other four word embedding models. It was also observed that the variation of Dunn's Index across different text corpora was higher for word2vec (skip-gram) embeddings as compared to other models, showing that the quality of clustering changed a lot more with changes in the training corpora. Word2vec (CBOW) embeddings portrayed a lower quality of clusters, indicating that embeddings using word2vec (CBOW) model formed word associations that were less distinctive than those formed using word2vec (skip-gram) model.

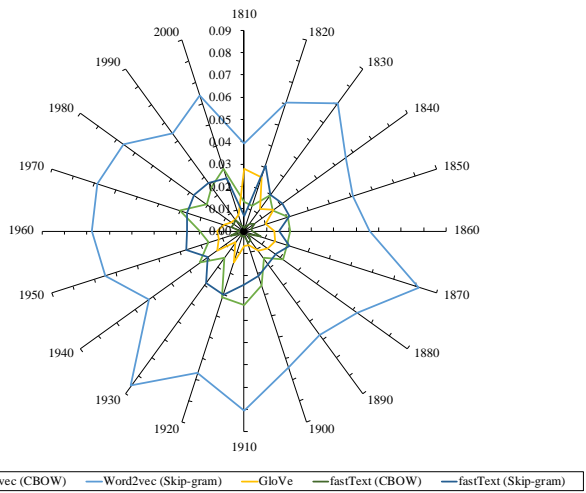

*Fig. 1. Variation in Dunn's Index across word embedding models*

This distinction between clustering done using CBOW and skip-gram embedding architectures was also visible for fastText word embedding models. On an average and across all decades, fastText (skip-gram) models captured more distinctive word associations as compared to fastText (CBOW) models. Hence, for both word2vec and fastText, skip-gram architecture captured the similarities and dissimilarities between words more often than the CBOW architecture.

When considering the two architectures, CBOW and skip-gram, word embedding models with the skip-gram architecture created more distinctive clusters, indicating that they captured more information about the distinctions between different words, as compared to CBOW. Moreover, it can be seen that skip-gram architecture was affected more often by changes in the training text corpora, as compared to CBOW, which was reflected by the higher standard deviations for these models. This shows that the sensitivity of skip-gram models towards lexical regularities was higher than the CBOW models.

TABLE II. DUNN'S INDEX FOR WORD EMBEDDING MODEL CLUSTERS

| Model | $M^a$ (DI) | $SD^a$ (DI) |
|---|---|---|
| word2vec (CBOW) | 0.021 | 0.006 |
| word2vec (skip-gram) | 0.063 | 0.011 |
| GloVe | 0.012 | 0.006 |
| fastText (CBOW) | 0.004 | 0.002 |
| fasttext (skip-gram) | 0.023 | 0.006 |

a. M and SD are used to represent mean and standard deviation respectively

*2) Distribution of Words across Clusters*

The distribution of words across each cluster for all five word embedding models, is shown in TABLE III. It should be noted that if words are uniformly distributed within clusters, each cluster would contain a similar proportion of words. This would lead to cluster populations deviating within a narrower range. If words are unevenly distributed within clusters, some clusters would contain a much higher proportion of words than other clusters, leading to cluster populations deviating across a wider range.

It can be seen from TABLE III that word2vec (skip-gram) and fastText (CBOW & skip-gram) models created more uniform clusters as compared to word2vec (CBOW) and GloVe models. Overall, the most uniform clustering was achieved by fastText (skip-gram) models, followed by fastText (CBOW) and word2vec (skip-gram).

TABLE III. DISTRIBUTION OF WORDS ACROSS EMBEDDING CLUSTERS

| % of words/cluster | Word2vec (CBOW) | Word2vec (SG) | GloVe | fastText (CBOW) | fastText (SG) |
|---|---|---|---|---|---|
| Min. | 0.12% | 1.14% | 0.03% | 0.82% | 1.77% |
| Mean | 12.50% | 12.50% | 12.50% | 12.61% | 12.53% |
| Max. | 82.34% | 51.17% | 79.81% | 44.52% | 34.14% |

For word2vec (CBOW) and GloVe models, some clusters contained more than 75% of the words, grouping most of the words together. This shows a lack of distinctive word associations for GloVe and word2vec (CBOW) word embedding models, as compared to other word embedding models. When comparing the five words embedding models, fastText word embeddings tend to create the most uniform clusters over the same training corpora. This is followed by word2vec embeddings, which created more uniform clusters as compared to GloVe embeddings. When comparing the CBOW and skip-gram architectures, it was revealed that for both word2vec and fastText models, the skip-gram architecture led to more uniform clustering.

Thus, even though a given text corpus might contain a common set of themes, these themes are captured differently across the five words embedding models. This again reveals that the word associations encoded by the word embeddings differ across the different word embedding models, which capture different lexical regularities from the same corpus.

*3) Jaccard Similarity*

The average Jaccard Similarity between clusters created for each pair of word embedding models is reported in TABLE III. These results show that word2vec (CBOW) and GloVe embeddings captured more similar word associations, as compared to word2vec (CBOW) and word2vec (skip-gram) embeddings. Hence, even though both the word2vec models had a common architecture, the difference in considering the word context to create the embeddings changed the way word associations were captured.

Moreover, word2vec (skip-gram) and fastText (skip-gram & CBOW) embeddings created very similar clusters as compared to word2vec (skip-gram) and word2vec (CBOW) embeddings. FastText (CBOW) and fastText (skip-gram) models created the most similar clusters, across pairs of all five word embedding models. The distinction between CBOW and skip-gram architecture was more visible in the clustering patterns of word2vec as compared to fastText. This signifies that word associations captured by word2vec models are more affected by the network architecture used to derive the embedding, i.e. whether the word has been predicted given a context, versus the context is predicted given a word, as compared to fastText models.

TABLE IV. JACCARD SIMILARITY BETWEEN CLUSTERS

| Model | word2vec[a] | word2vec[b] | GloVe | fastText[a] | fastText[b] |
|---|---|---|---|---|---|
| word2vec[a] | 1 | - | - | - | - |
| word2vec[b] | 0.38 | 1 | - | - | - |
| GloVe | 0.44 | 0.36 | 1 | - | - |
| fastText[a] | 0.34 | 0.46 | 0.35 | 1 | - |
| fastText[b] | 0.39 | 0.5 | 0.41 | 0.62 | 1.00 |

a. CBOW
b. Skip-gram

## B. Variation in word associations over varying text corpora

### 1) Variation in Dunn's Index

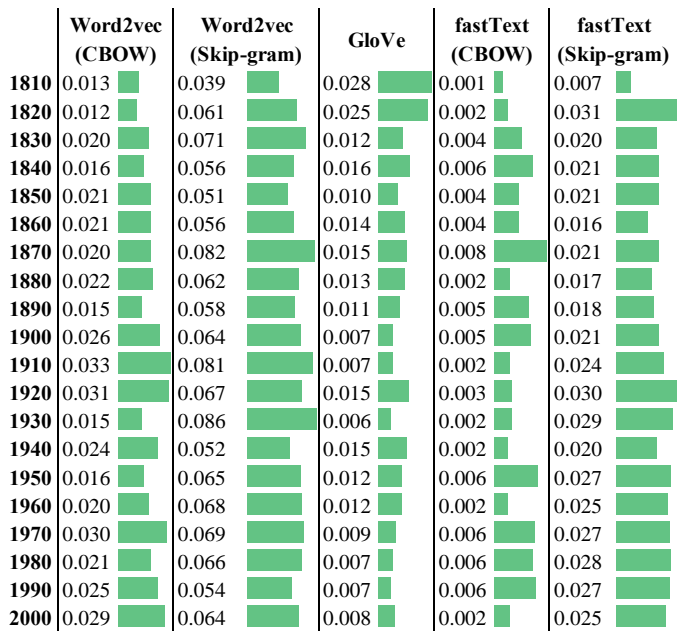

| | Word2vec (CBOW) | Word2vec (Skip-gram) | GloVe | fastText (CBOW) | fastText (Skip-gram) |
|---|---|---|---|---|---|
| 1810 | 0.013 | 0.039 | 0.028 | 0.001 | 0.007 |
| 1820 | 0.012 | 0.061 | 0.025 | 0.002 | 0.031 |
| 1830 | 0.020 | 0.071 | 0.012 | 0.004 | 0.020 |
| 1840 | 0.016 | 0.056 | 0.016 | 0.006 | 0.021 |
| 1850 | 0.021 | 0.051 | 0.010 | 0.004 | 0.021 |
| 1860 | 0.021 | 0.056 | 0.014 | 0.004 | 0.016 |
| 1870 | 0.020 | 0.082 | 0.015 | 0.008 | 0.021 |
| 1880 | 0.022 | 0.062 | 0.013 | 0.002 | 0.017 |
| 1890 | 0.015 | 0.058 | 0.011 | 0.005 | 0.018 |
| 1900 | 0.026 | 0.064 | 0.007 | 0.005 | 0.021 |
| 1910 | 0.033 | 0.081 | 0.007 | 0.002 | 0.024 |
| 1920 | 0.031 | 0.067 | 0.015 | 0.003 | 0.030 |
| 1930 | 0.015 | 0.086 | 0.006 | 0.002 | 0.029 |
| 1940 | 0.024 | 0.052 | 0.015 | 0.002 | 0.020 |
| 1950 | 0.016 | 0.065 | 0.012 | 0.006 | 0.027 |
| 1960 | 0.020 | 0.068 | 0.012 | 0.002 | 0.025 |
| 1970 | 0.030 | 0.069 | 0.009 | 0.006 | 0.027 |
| 1980 | 0.021 | 0.066 | 0.007 | 0.006 | 0.028 |
| 1990 | 0.025 | 0.054 | 0.007 | 0.006 | 0.027 |
| 2000 | 0.029 | 0.064 | 0.008 | 0.002 | 0.025 |

*Fig. 2. Variation in Dunn's index across varying text corpora (1810s - 2000s)*

Over different decades, the quality of clusters, as measured by Dunn's Index, varied for the given five word embedding models, even when trained on the same training corpus. Fig. 2 shows this variation over different decades as well as different word embedding models. Across all five word embedding models, it can be seen that word2vec (skip-gram) model architecture outperformed the other four models, in terms of capturing distinctive word associations from a given text. The clustering quality also improved for all of the models as the size of the corpus increased, apart from GloVe, whose clustering quality degraded as the text corpus became larger.

Overall, it can be observed that the skip-gram architecture created better quality clusters, i.e. skip-gram word embeddings were able to encode more distinctive word associations, capturing a higher degree of similarity and dissimilarity between words within the vector space, as compared to the CBOW word embeddings.

## V. CONCLUSION

This paper proposed a semi-supervised analysis approach for attributive word associations developed in five different word embedding models, trained over 20 decades of text data. Quantitative analysis of attributive word associations revealed significant changes in the quality and strength of word associations with respect to changes in word embedding models and training corpora. We demonstrate that the context learning architecture, type of embedding model and size of training corpora influence the embedding spaces generated. The choice of context learning architecture influences how sensitive word embeddings are towards changes in training corpora, skip-gram being more sensitive to changes in the training corpora as compared to CBOW. The skip-gram architecture captures more distinguishable word associations as compared to CBOW, i.e. it encodes embedding spaces where words deemed similar by the embedding model lie closer together and dissimilar words lie farther apart, increasing distinguishability in word associations. Across embedding models, word2vec encodes the most distinguishable word associations as compared to fastText and GloVe. The size of the training corpora also affects the quality of word associations generated. Moreover, the distinguishability of word associations generated increased when the models are trained on larger corpora, with the exception of GloVe, where word association strength degraded possibly due to the limited number of training epochs.

Our future work on analyzing attributive word associations will focus on creating a corpus specific taxonomy of adjectives, assisting in determining the baseline for clusters created, improving the accuracy and specificity of the analyses. More research is required on model hyperparameter selection for word embedding models to ensure inter-model comparability. It is also crucial to account for the vast variations in the vocabulary from 1810s to 2000s, which ranges from rare archaic English words to recently created slang words. This will also improve the accuracy of the words being tagged and the embeddings being generated by capturing the context in which each word is used.